\documentclass[runningheads,a4paper]{llncs}
\usepackage{makeidx}  
\usepackage{graphicx}
\usepackage{mathtools}
\usepackage{color}
\usepackage{amssymb}
\usepackage{amsmath}
\usepackage{graphicx}
\usepackage{epstopdf}
\usepackage{supertabular}
\usepackage{array}
\usepackage{multirow}
\usepackage{setspace}
\usepackage{booktabs}
\usepackage{hyperref} 

\usepackage[caption=false,font=footnotesize]{subfig}

\usepackage[caption=false,font=footnotesize]{subfig}
\newcommand{\keywords}[1]{\par\addvspace\baselineskip
	\noindent\keywordname\enspace\ignorespaces#1}

\begin{document}
\mainmatter              
\title{Just-Enough Interaction Approach to Knee MRI Segmentation: Data from the Osteoarthritis Initiative}
\titlerunning{JEI Approach to Knee MRI Segmentation}

\author{Satyananda Kashyap \and Honghai Zhang \and Milan Sonka}
\authorrunning{Kashyap et. al}
\tocauthor{Satyananda Kashyap,Honghai Zhang,Milan Sonka}

\index{Kashyap, Satyananda} 
\index{Zhang,Honghai}
\index{Sonka,Milan}

\institute{Iowa Institute for Biomedical Imaging, The University of Iowa, Iowa City IA, USA\\
\email{satyananda-kashyap@uiowa.edu}}

\maketitle              
\begin{abstract}

State-of-the-art automated segmentation algorithms are not 100\% accurate especially when segmenting difficult to interpret datasets like those with severe osteoarthritis (OA). We present a novel interactive method called just-enough interaction (JEI), which adds a fast correction step to the automated layered optimal graph segmentation of multiple objects and surfaces (LOGISMOS). After LOGISMOS segmentation in knee MRI, the JEI user interaction does not modify boundary surfaces of the bones and cartilages directly. Local costs of underlying graph nodes are modified instead and the graph is re-optimized, providing globally optimal corrected results. Significant performance improvement ($p \ll 0.001$) was observed when comparing JEI-corrected results to the automated. The algorithm was extended from 3D JEI to longitudinal multi-3D (4D) JEI allowing simultaneous visualization and interaction of multiple-time points of the same patient.

\end{abstract}

\section{Introduction}
Osteoarthritis (OA) is one of the most widely afflicted diseases among the aging population with no interventional drugs available \cite{jama1990}. The current standard care for the disease is prescription of analgesics and using joint braces to ease the pain with worsening of OA eventually leading to total knee replacement. MRI as a non-invasive bio-marker has potential to detect the structural changes early to help predict the onset of OA and monitor the effects of treatments in drug trials. For any analysis the crucial first step is the segmentation of the bones and cartilages of the femur and tibia. Manual segmentation of a MR volume takes several hours and is subjected to inter and intra observer biases. Several automated algorithms exist, which are robust for most of the cases but cannot be relied upon for subjects with severe pathology. With disease progression, automated algorithms face a challenging problem of delineating the bones and cartilages in presence of bone marrow lesions, cartilage surface thinning, mesical extrusion and synovial fluid leakage. Many of these artifacts are symptoms of the disease and appear similar in texture and intensity to cartilage on MR volumes. 

Interactive correction methods are designed to help ease the post-processing needed. Several such techniques have been proposed in the literature such as thin-plate splines \cite{ross2009lung}, active shape models for interactions \cite{schwarz2008interactive} and live-wires \cite{schenk2000efficient}. Live-wires have embedded user interactions that require a seed based initialization which is user inputted or based on some initialization algorithm. However they are inherently 2D mechanisms which were later extended to 3D with a drawback of being unable to maintain global optimality for multiple surfaces and objects. Several of the above mentioned methods correct for segmentation inaccuracies by directly matching the object boundaries with the interaction which after several repetitions results in the final surfaces having local topological errors. 

The proposed interaction algorithm uses a graph based LOGISMOS framework \cite{Li2006PAMI,Yin2010} with the user-clicked points hereafter called \emph{nudge points} interacting directly with the underlying graph framework. This method has guarantees of global optimality for every interaction and differs from the traditional voxel-by-voxel editing by requiring just enough (i.e., limited) interaction (JEI) to correct the original automated segmentation if needed. The proposed method may appear similar to the Boykov's graph cut \cite{boykov2001interactive}, however their interaction algorithm is not able to guarantee global optimality for multiple surfaces and objects. LOGISMOS-JEI always guarantees global optimality when handling multiple objects and surfaces. The JEI architecture and GUI are designed to be platform and application agnostic and their details are covered in \cite{Honghai-2016-MICCAI-Workshop}.
The algorithm is extended to handle longitudinal JEI (multi-3D, or 4D) which enables correction of multiple time-points of the single patient sequence. It is based on an existing automated 4D LOGISMOS framework \cite{Kashyap2015}. The GUI was also extended to enable simultaneous examination and interaction using all the time-points giving a progressive view of cartilage losses.

\section{Methods}
The JEI method starts with an initial automated LOGISMOS segmentation. Post segmentation, the resulting optimized graph state (called residual graph) is saved for the purposes of JEI. The JEI algorithm was extended from \cite{sun2013} to handle multi-surface multiple object interactions. Further, a new interaction mechanism was developed along with a faster graph optimization library to provide almost immediate feedback on the interaction. 

\subsection{Automated LOGISMOS algorithm}
The automated LOGISMOS algorithm was detailed in \cite{Yin2010}. The algorithm identified the volumes of interest (VOI's) which pertain to a smaller regions around the femur and tibia. This VOI identification was done using an Adaboost classifier which was trained on example VOI's using 3D Haar-like features. The VOI detection helped reduce the computation time by operating on the smaller region. Further the VOI bounds were used for affine fitting the mean shape mesh $S_0$ for the femur and tibia bone respectively. Since a patient specific bone shape affects the final segmentation, a single surface single object LOGISMOS segmentation was computed using $S_0$. 

Each surface of the object to be segmented had a geometric graph constructed. A set of graph nodes were built up as a column from the patient specific mesh surface $S$ which represented the set of locations for the final surfaces. Every node was associated with either a bone or a cartilage surface cost. The bone costs along the column direction were 1D derivative operator $\nabla(x,y,z)$. Similarly the cartilage costs was an empirically weighted ($w$) combination of first and second order derivatives given by $ w * \nabla(x,y,z) + (1-w)* \nabla^2(x,y,z)$. To ensure topologically correct segmentation non-intersecting geometric graph  construction is crucial. This was enforced by replicating the column construction as non-intersecting electric lines of force with the mesh surface points denoting the positively charged particles. The surface segmentation task was solved by representing the problem as a max-flow optimization which was accomplished by connecting intra and inter-columns arcs to enforce the smoothness constraints. Further inter-object and inter-surface constraints were constructed for the final multi-surface multi-object segmentation.

For the 4D automated multi-object multi-surface multi-3D segmentation \cite{Kashyap2015}, inter-time-point graph arcs were introduced which helped establish the maximum and minimum allowable changes between the time-points to be within physiologically permissible ranges. The inter-time point arcs longitudinally linked the corresponding column positions of the bones and the cartilages of the femur and tibia temporally. 


\subsection{JEI Work-flow}
\begin{figure}[htb]
	\centering
	\includegraphics[width=0.9\textwidth]{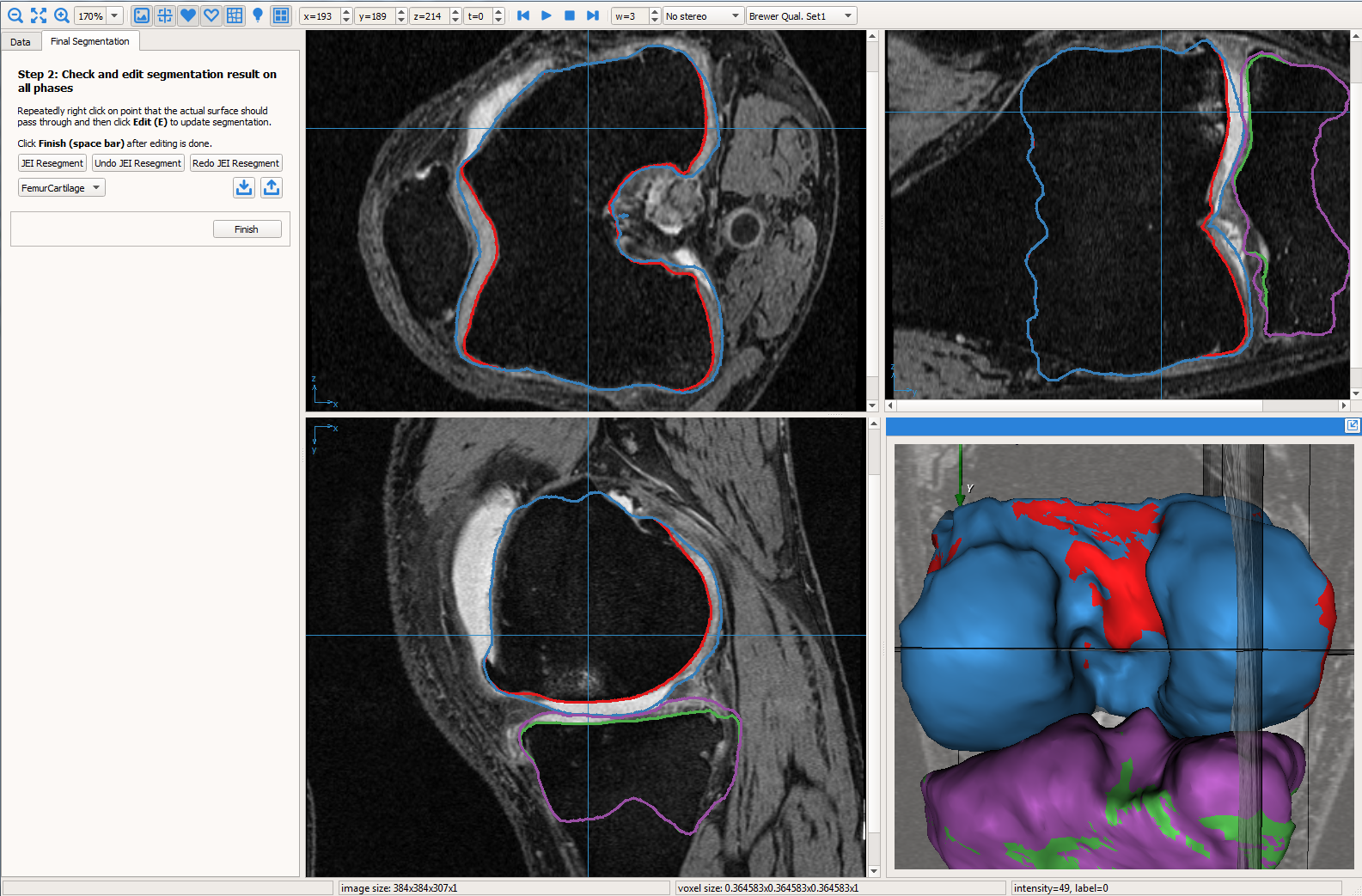}
	\caption{The graphical user interface for 3D JEI with the image volumes and the surface meshes overlayed.}
	\label{fig:Interface}
\end{figure}

\begin{figure}[htb]
	\centering
	(a) \includegraphics[width=0.8\textwidth]{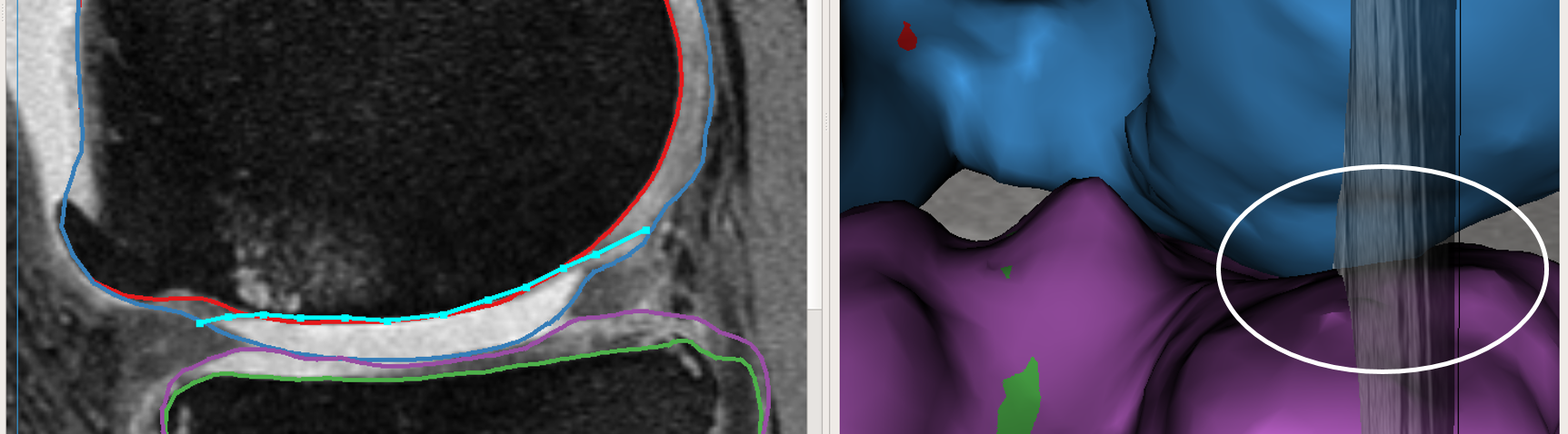}\\
	(b) \includegraphics[width=0.8\textwidth]{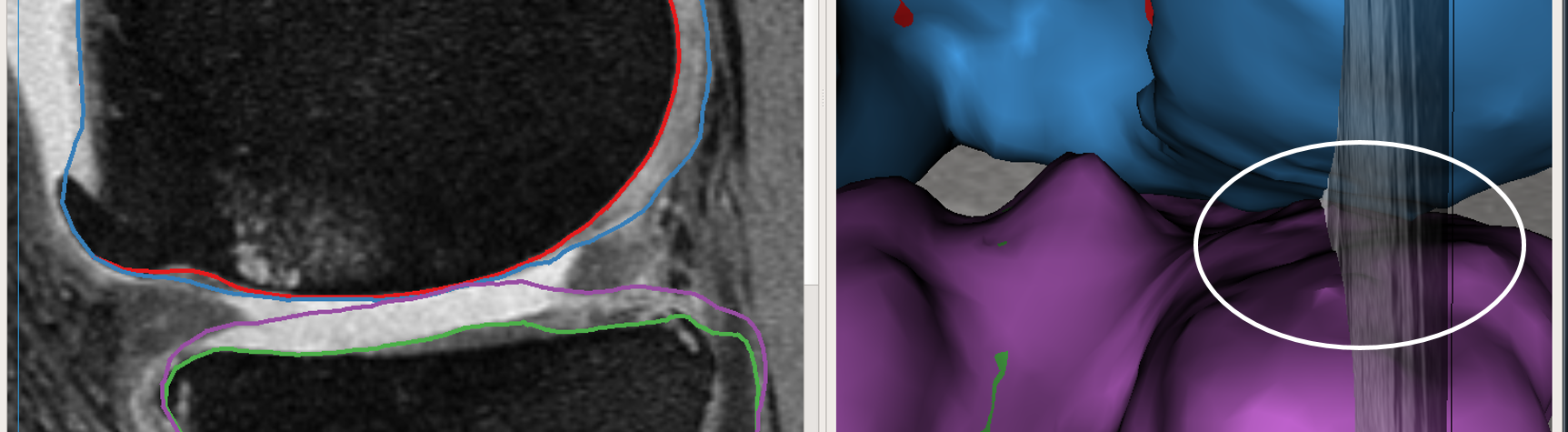}\\
	(c) \includegraphics[width=0.8\textwidth]{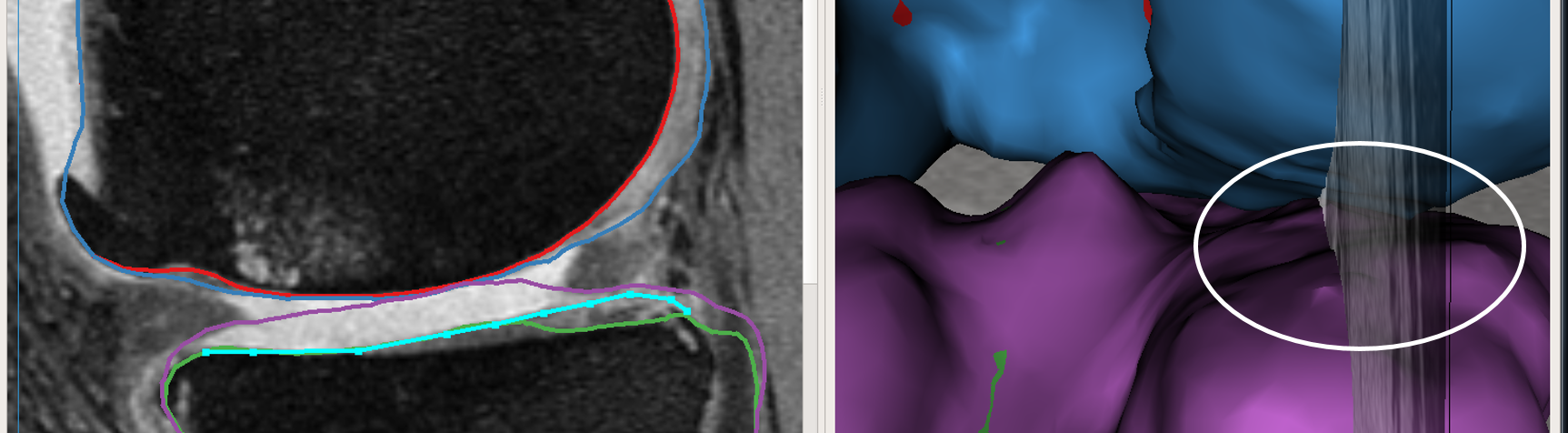}\\
	(d) \includegraphics[width=0.8\textwidth]{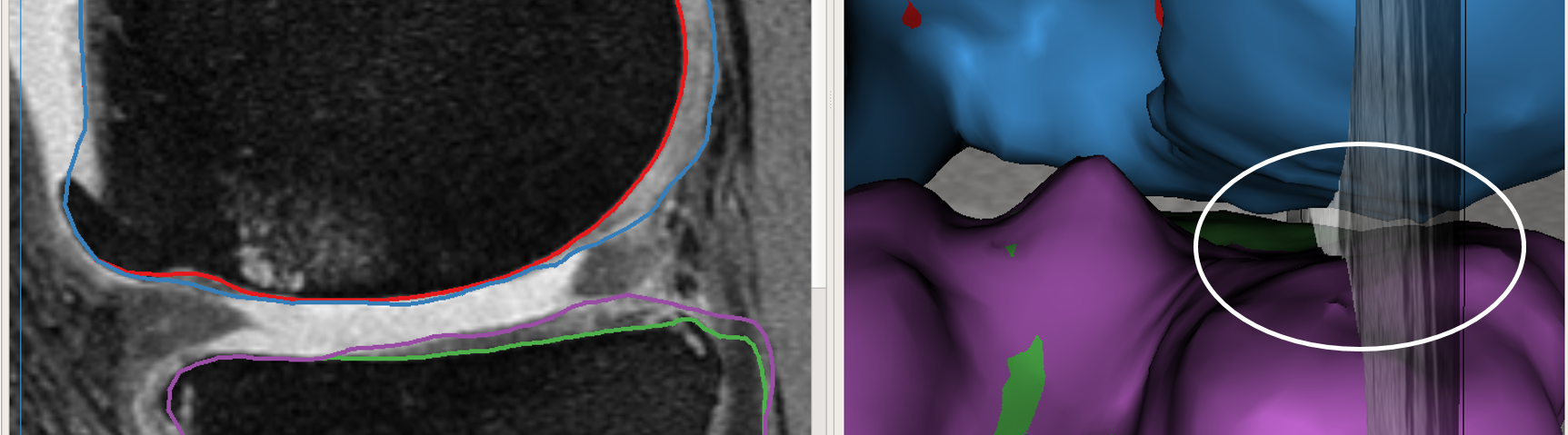}
	\caption{JEI work-flow to correct segmentation inaccuracy. The circled region indicates the 3D neighborhood correction based on a single 2D slice editing. The four surfaces shown as contours and 3D objects are the femur bone and cartilage (colored red and blue respectively) and the tibia bone and cartilage (colored green and purple respectively). The nudge points marked in the intermediate steps are shown in cyan. A detailed video demonstration of the interaction steps is available at \href{http://bit.ly/2blYXFz}{http://bit.ly/2blYXFz}}
	\label{fig:Workflow}
\end{figure}

The electric lines of force (ELF) based geometric graph, image volume and residual graph were loaded into the GUI (see Fig.~\ref{fig:Interface}) to inspect segmentation quality and perform JEI. The work-flow is presented as a video provided in the supplementary material with an example subject with severe OA. The details of the work-flow were as follows: 

\begin{enumerate}
	\item {\bf User provided nudge points:} The user identified correction is provided as a set of nudge points which guide the segmentation to the correct position. Fig.~\ref{fig:Workflow}a shows the GUI magnified with the volume and the automated LOGISMOS segmentation results overlayed. The particular slice indicated is a case with severe OA having bright fluid regions improperly segmented as cartilage. The blue line with points are the nudge points indicated by the user approximately identifying the correct cartilage region.
	\item {\bf 3D local graph cost modification:} To identify the underlying graph columns influenced by the nudge points (defined as a contour), a $k$-dimensional tree algorithm is used which stores all the geometric graphs positions. In previous JEI applications \cite{sun2013} the graph was constructed on a regular 3D grid where the nearest graph columns could be identified quite easily. However given the complex shape of the knee objects and the ELF graph constructed based on it, a more sophisticated query of the closest columns is needed which does not compromise on speed. The $k$D tree allows for a $O(log~n)$ query on the N nearest graph nodes (empirically determined) for every nudge point. Once identified the costs (i.e. unlikeliness) at corresponding columns associated with these nodes are modified as 
	\[
	c(i,j)= 
	\begin{dcases}
	0, & \text{if } D((i,j),n(i,j)) < \Delta\\
	1, & \text{otherwise}
	\end{dcases} \; ,
	\]
	with $c(i,j)$ defined as the cost of node $j$ on column $i$, $D((i,j),n(i,j))$ the distance between node closest to the nudge point $(i,j)$ and its nearby intersecting nodes $n(i,j)$ within the $\Delta$ tolerance.	
	\item {\bf Max-flow re-computation:} Following the local graph cost modification the max-flow is recomputed in 3D within a few milliseconds and the updated surfaces rendered onto the GUI. As seen in Fig.~\ref{fig:Workflow}b the correction made by the nudge points are reflected in the updated cartilage surface overlayed on the image volume. 
\end{enumerate}

The above work-flow is repeated to correct the tibial cartilage errors as well. In the intermediate steps following the correction of the femur, the tibia bone and cartilage surfaces appear to worsen. This can be attributed to a combination of the existing graph costs and the graph constraints. Since the tibia cartilage surface has no clear defined edge cost in that region, the surface result moved along with the femur corrected cartilage surface. Subsequently due to the inter-surface distance constraints between the tibial surfaces the tibial bone surface also changed. However once the nudge points provided the appropriate locations for cost modification the erroneous surfaces were corrected (Figs.~\ref{fig:Workflow}c,d). Note that the corrections made on a single 2D slice resulted in the entire locally affected 3D neighborhood being corrected. This can be appreciated in the corresponding circled regions of the surface model. 

\subsubsection{Undo-Redo Interaction Capabilities}
For a more consistent segmentation interaction we designed into the GUI the ability to save the user interaction steps. A stack was used to save the inputted nudge points and the surface ID for each editing step. The editing was reverted by popping the stack which restored the previous costs on the local graph columns. Re-optimizing the graph resulted in the previous surfaces. However the popped stack was not deleted unless a different interaction was continued after being reverted. If the interaction needed to be redone, the pointer simply moved to the previous position on the stack and repeated the same steps as above to redo the correction. 
 
When a new automated LOGISMOS surface is loaded that was previously edited the user can load the editing stack to bring the interaction to most up-to date edited state. The edits can be continued from that point. We anticipate that this feature would be very useful in reducing the inter-observer and intra-observer variability. A video demonstration \footnote{\href{ http://bit.ly/2blYXFz}{ http://bit.ly/2blYXFz}} of this feature is provided in the supplementary material. 

\subsection{Longitudinal JEI}
\begin{figure}[htb]
	\centering
	(a) \includegraphics*[width=0.9\columnwidth]{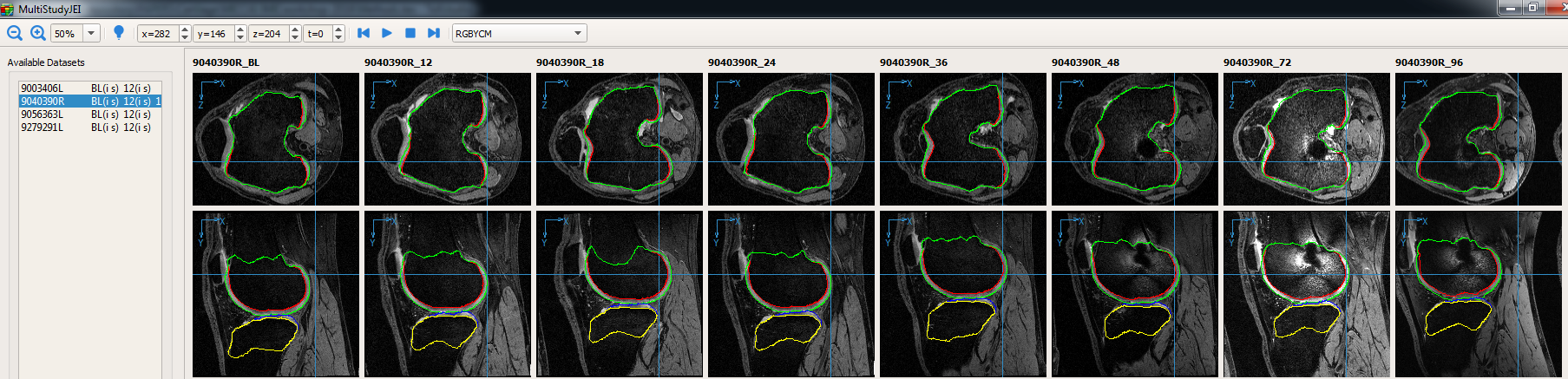} \\
	(b) \includegraphics*[width=0.7\columnwidth]{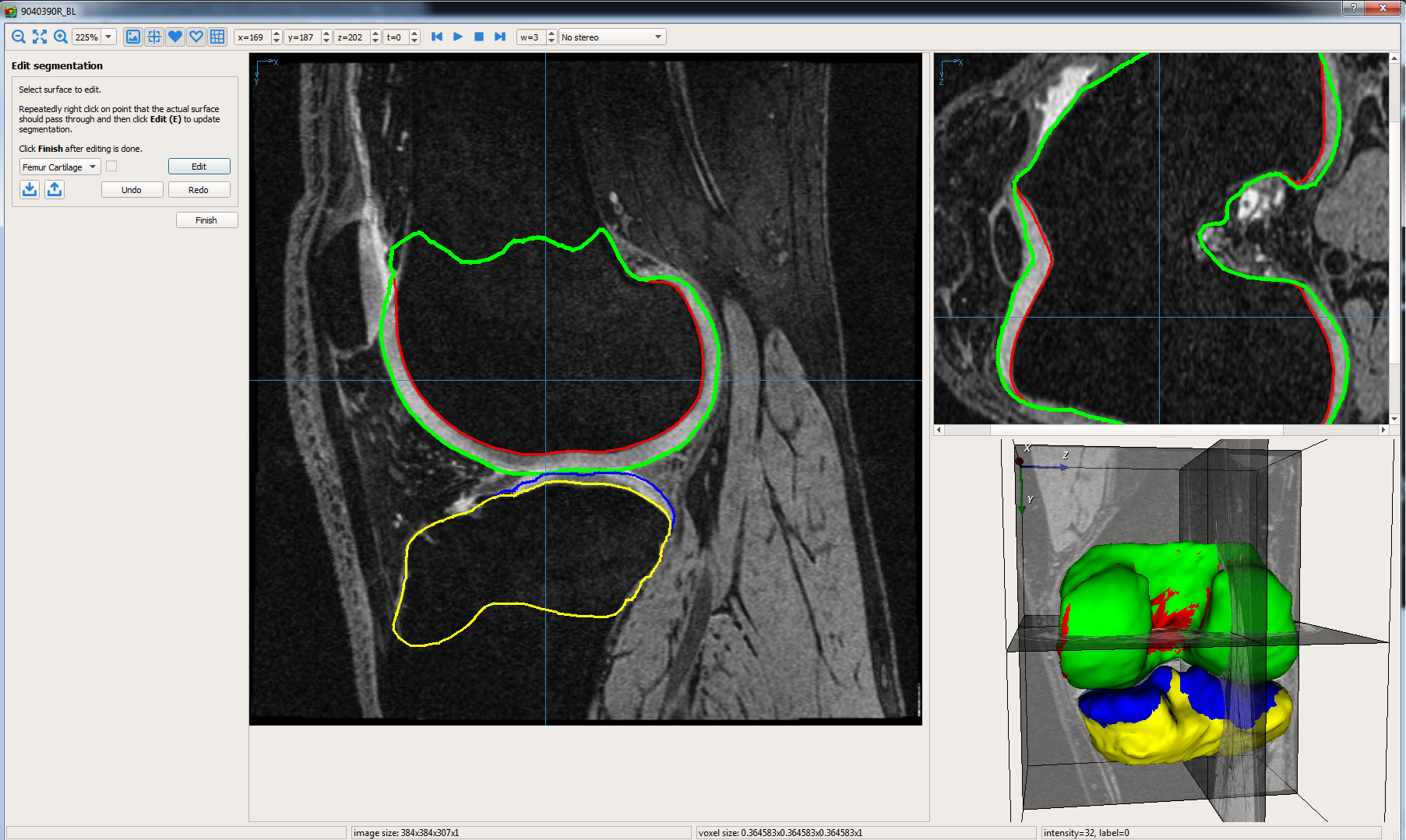}
	\caption{4D LOGISMOS-JEI. (a)~Longitudinal JEI viewer screen-shot showing a thumbnail of eight time-points of a single patient simultaneously. (b)~Smaller editing window for each 3D time-point.}
	\label{fig:multiTime}
\end{figure}

 The longitudinal JEI GUI was extended to enable visualization of all patient time-points simultaneously. The viewer also enabled synchronized scrolling across datasets. Fig.~\ref{fig:multiTime}a shows eight time-points of the same patient (baseline, 12, 18, 24, 36, 48, 72 and 96 month follow-ups) being simultaneously visualized. Each individual thumbnail view can be expanded (see Fig.~\ref{fig:multiTime}b) to a detailed larger GUI (identical to 3D GUI) for interaction. 
 
 The interaction mechanism is similar to the 3D JEI where a set of nudge points on a single 2D slice modifies the graph node costs in the local 3D neighborhood columns of the given time-point. Further since the longitudinal JEI has has a single large underlying residual graph with temporal inter-time-point constraints the corresponding local 3D neighborhood column locations at the other time-points are also corrected.
 
\section{Experimental Methods}
MRI volumes used in this study were acquired from the osteoarthritis initiative which also had a limited number of datasets with independent standard available. 
All subjects were scanned using the DESS protocol with a voxel resolution of $0.36\times0.36\times0.7$ mm$^3$. 19 baseline subjects with varying degrees of OA severity were used in this study. They were segmented using the automated LOGISMOS followed by 3D JEI correction. The geometric graph for the tibia and femur objects had 8006 and 8002 graph columns, respectively. The graph parameters used in this experiment are listed in Table~\ref{graph-parameters}.

\begin{table}[htb]
	\centering
	\caption{Parameters used for graph construction. Minimum inter-surface inter-object and inter-time-point separations are zero. Note that the inter-time point constraints were only used in the longitudinal JEI.}
	\label{graph-parameters}
	\begin{tabular}{lcccccc}
		\hline
		& Inter-surface & Inter-object &Inter-time point& Smoothness & Column size & Node spacing \\
		& max (nodes)   & max (nodes)& max (nodes) & (nodes)    & (nodes)     & (mm) \\
		\hline
	    & 20            & 60&  5  & 2          & 61          & 0.20 \\
		\hline
	\end{tabular}
\end{table}

%
\section{Results}
The surface positioning errors of the automated LOGISMOS and the JEI corrected surfaces were computed against the independent standard which were manually traced cartilage and bone borders of the knee joint provided by the OAI. The bone surface segmentation results were very robust. The presented results focused on the cartilage surface. Table~\ref{error-table} shows the 3D JEI versus automated LOGISMOS for 19 baseline datasets. The signed errors for the JEI-corrected surfaces were close to zero and were significantly smaller ($p\ll 0.001$) for the femur cartilage. Although the signed tibia error appeared to have a lower mean for automated LOGISMOS, the difference was not statistically significant.
The unsigned errors for the JEI-corrected surfaces were significantly smaller ($p\ll 0.001$) for both the femur and tibia. After JEI, the errors decreased to about a half of the original LOGISMOS surface errors.

\begin{table}[]
	\centering
	\caption{Surface positioning errors (signed, unsigned, in mm) of 3D JEI-corrected versus automated LOGISMOS segmentation. Bold entries mark statistical significantly better performance of the pairwise comparisons.}
	\label{error-table}
	\resizebox{\textwidth}{!}{%
		\begin{tabular}{@{}|l|l|l|l|l|l|l|l|@{}}
			\toprule
			& JEI-Corrected             & Automated          & p-value  &                & JEI-Corrected            & Automated  & p-value  \\ \midrule
			Femur Signed & {\bf -0.03 $\pm$ 0.08} & -0.37 $\pm$ 0.23        & $\ll0.001$ & Femur Unsigned & {\bf 0.43 $\pm$ 0.06} & 0.73 $\pm$ 0.16 & $\ll0.001$ \\ \midrule
			Tibia Signed & 0.03 $\pm$ 0.17          & 0.01 $\pm$ 0.49 & 0.89     & Tibia Unsigned & {\bf 0.51 $\pm$ 0.13} & 0.80 $\pm$ 0.20 & $\ll0.001$ \\ \bottomrule
		\end{tabular}%
	}
\end{table}

\section{Discussion \& Conclusions }
3D JEI and longitudinal JEI methods for knee MRI were presented. The interactive corrections for 3D JEI took on an average 15 min per dataset in comparison to several hours of effort that are needed for traditional voxel-by-voxel editing. The surface positioning errors of JEI-corrected results showed significant improvements over the automated LOGISMOS. 

Another example application, not reported above, used the JEI-corrected surfaces to train machine learning classifiers to identify cartilage regions in the presence of pathology.  For the fully automated LOGISMOS, using JEI-trained classifiers yielded significantly better performance than employing simple cost functions. The learning based method will be presented in \cite{Kashyap2016-MICCAI}. Further quantitative analysis of longitudinal JEI, comparing, and quantifying the inter/intra observer bias and validating JEI corrected results in a larger clinical study are planned for future work. 

%


%


\section*{Acknowledgments}
This research was supported by NIH grant R01-EB-004640. The OAI is a public-private partnership comprised of five contracts (N01-AR-2-2258; N01-AR-2-2259; N01-AR-2-2260; N01-AR-2-2261; N01-AR-2-2262) funded by the National Institutes of Health, a branch of the Department of Health and Human Services, and conducted by the OAI Study Investigators. Private funding partners include Merck Research Laboratories; Novartis Pharmaceuticals Corporation, GlaxoSmithKline; and Pfizer, Inc. Private sector funding for the OAI is managed by the Foundation for the National Institutes of Health. This manuscript was prepared using an OAI public use data set and does not necessarily reflect the opinions or views of the OAI investigators, the NIH, or the private funding partners
{\small
\bibliographystyle{splncs}
\bibliography{MICCAI-IMICworkshop2016}
}
\end{document}